\documentclass[10pt,twocolumn,letterpaper]{article}

\usepackage{faceformer}              

\usepackage{graphicx}
\usepackage{amsmath}
\usepackage{amssymb}
\usepackage{booktabs}

\usepackage[pagebackref,breaklinks,colorlinks]{hyperref}

\usepackage[capitalize]{cleveref}
\crefname{section}{Sec.}{Secs.}
\Crefname{section}{Section}{Sections}
\Crefname{table}{Table}{Tables}
\crefname{table}{Tab.}{Tabs.}

\begin{document}
\title{FaceFormer: Speech-Driven 3D Facial Animation with Transformers}

\author{Yingruo Fan$^1$ \quad Zhaojiang Lin$^{2\dagger}$ \quad Jun Saito$^3$ \quad Wenping Wang$^{1,4}$ \quad Taku Komura$^{1\ast}$\\[0.3em]
$^1$The University of Hong Kong \quad $^2$The Hong Kong University of Science and Technology\\
\quad $^3$Adobe Research \quad $^4$Texas A\&M University 
}

\maketitle

\newcommand\blfootnote[1]{%
\begingroup
\renewcommand\thefootnote{}\footnote{#1}%
\addtocounter{footnote}{-1}%
\endgroup
}

\blfootnote{$\ast$ Corresponding author}
\blfootnote{$\dagger$ Work done at HKUST}


\begin{abstract}
Speech-driven 3D facial animation is challenging due to the complex geometry of human faces and the limited availability of 3D audio-visual data. Prior works typically focus on learning phoneme-level features of short audio windows with limited context, occasionally resulting in inaccurate lip movements. To tackle this limitation, we propose a Transformer-based autoregressive model, FaceFormer, which encodes the long-term audio context and autoregressively predicts a sequence of animated 3D face meshes. To cope with the data scarcity issue, we integrate the self-supervised pre-trained speech representations. Also, we devise two biased attention mechanisms well suited to this specific task, including the biased cross-modal multi-head (MH) attention and the biased causal MH self-attention with a periodic positional encoding strategy. The former effectively aligns the audio-motion modalities, whereas the latter offers abilities to generalize to longer audio sequences. Extensive experiments and a perceptual user study show that our approach outperforms the existing state-of-the-arts. We encourage watching the video\footnote{The supplementary video and code are available at: \url{https://evelynfan.github.io/audio2face/}}.
\end{abstract}

\section{Introduction}

Speech-driven 3D facial animation has become an increasingly attractive research area in both academia and industry. It is potentially beneficial to a broad range of applications such as virtual reality, film production, games and education. Realistic speech-driven 3D facial animation aims to automatically animate vivid facial expressions of the 3D avatar from an arbitrary speech signal. 

\begin{figure}
\centering
\includegraphics[width=0.42\textwidth]{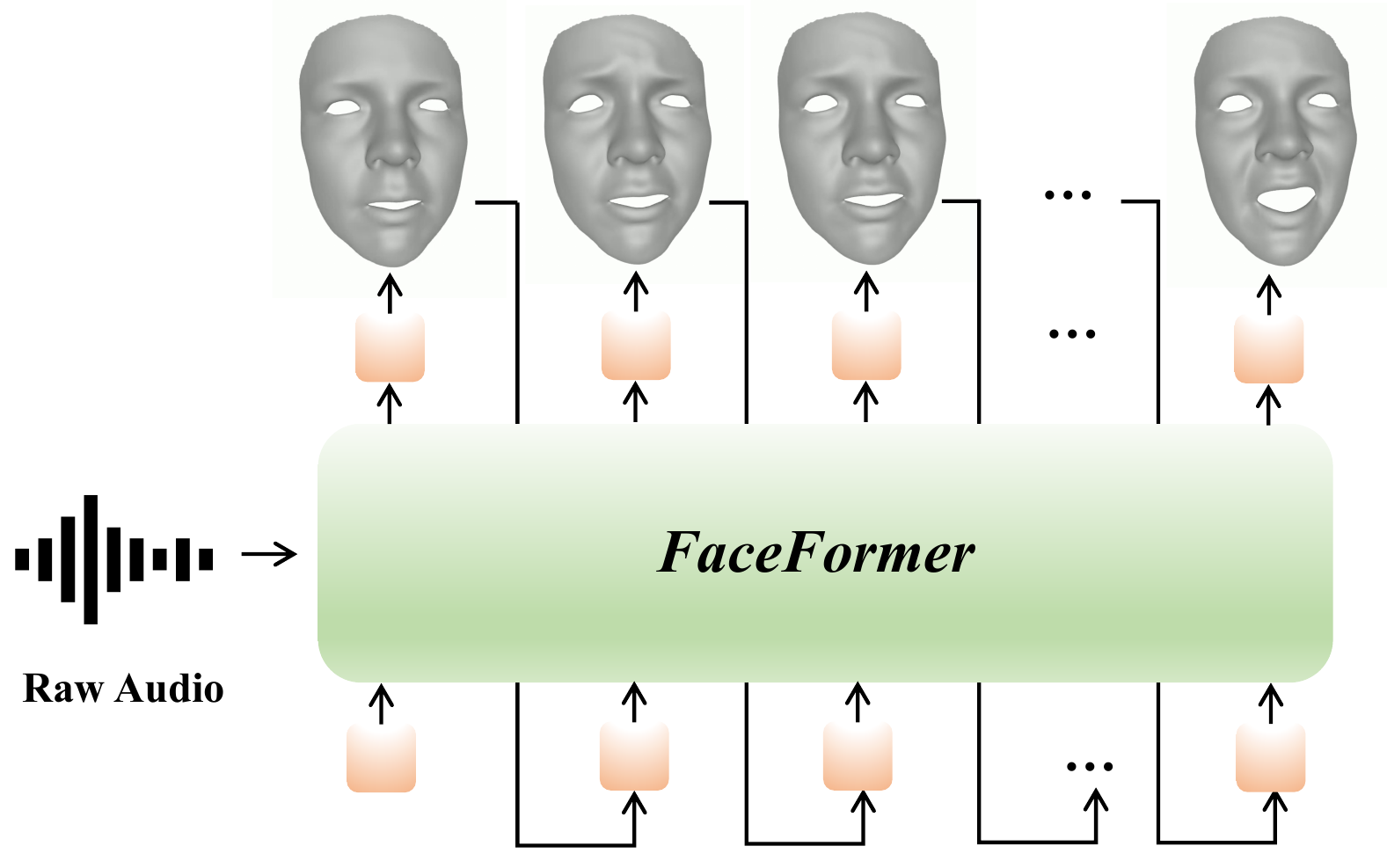}
\caption{\label{fig:illustration} \textbf{Concept diagram of FaceFormer.} Given the raw audio input and a neutral 3D face mesh, our proposed end-to-end Transformer-based architecture, dubbed FaceFormer, can autoregressively synthesize a sequence of realistic 3D facial motions with accurate lip movements. }
\end{figure}

We focus on animating the 3D geometry rather than the 2D pixel values, \eg photorealistic talking-head animation~\cite{chung2017you,suwajanakorn2017synthesizing,wiles2018x2face,chen2019hierarchical,zhou2019talking,mittal2020animating,zeng2020talking}. The majority of existing works aim to produce 2D videos of talking heads, given the availability of massive 2D video datasets. However, the generated 2D videos are not directly applicable to applications like 3D games and VR, which need to animate 3D models in a 3D environment. Several methods~\cite{pham2017speech,habibie2021learning,wang20213d} harness 2D monocular videos to obtain 3D facial parameters, which might lead to unreliable results. This is because the quality of the synthetic 3D data is bounded by the accuracy of 3D reconstruction techniques, which cannot capture the subtle changes in 3D. In speech-driven 3D facial animation, most 3D mesh-based works~\cite{cudeiro2019capture,chaispeech,liu2021geometry} formulate the input as short audio windows, which might result in ambiguities in variations of facial expressions. As pointed out by Karras~\etal~\cite{karras2017audio}, a longer-term audio context is required for realistically animating the whole face. While MeshTalk~\cite{richard2021meshtalk} has considered a longer audio context by modeling the audio sequence, training the model with Mel spectral audio features fails to synthesize accurate lip motions in data-scarce settings. 
Collecting 3D motion capture data is also considerably expensive and time-consuming. 

To address the issues about long-term context and lack of 3D audio-visual data, we propose a transformer-based autoregressive model (\cref{fig:illustration}) which (1) captures longer-term audio context to enable highly realistic animation of the entire face, \ie both upper and lower face expressions, (2) effectively utilizes the self-supervised pre-trained speech representations to handle the data scarcity issue, and (3) considers the history of face motions for producing temporally stable facial animation. 

Transformer~\cite{vaswani2017attention} has achieved remarkable performance in both natural language processing~\cite{devlin2018bert,vaswani2017attention} and computer vision~\cite{chen2020generative,parmar2018image,dosovitskiy2020image} tasks. The sequential models like LSTM have a bottleneck that hinders the ability to learn longer-term context effectively~\cite{pascanu2013difficulty}. Compared to RNN-based models, transformer can better capture long-range context dependencies based solely on attention mechanisms~\cite{vaswani2017attention}. Recently, transformer has also made the encouraging progress in body motion synthesis~\cite{aksan2020spatio,petrovich2021action,bhattacharya2021text2gestures} and dance generation~\cite{li2020learning,li2021AIST++,valle2021transflower}. The success of transformer is mainly attributed to its design incorporating the self-attention mechanism, which is effective in modeling both the short- and long-range relations by explicitly attending to all parts of the representation. Speech-driven 3D facial animation has not been explored in this direction. 

Direct application of a vanilla transformer architecture to audio sequences does not perform well on the task of speech-driven 3D facial animation, and we thus need to address these issues.  First, transformer is data-hungry in nature, requiring sufficiently large datasets for training~\cite{khan2021transformers}. Given the limited availability of 3D audio-visual data, we explore the use of the self-supervised pre-trained speech model wav2vec 2.0~\cite{baevski2020wav2vec}. Wav2vec 2.0 has learned rich phoneme information, since it has been trained on a large-scale corpus~\cite{panayotov2015librispeech} of unlabeled speech. While the limited 3D audio-visual data might not cover enough phonemes, we expect the pre-trained speech representations can benefit the speech-driven 3D facial animation task in data-scarce settings. Second, the default encoder-decoder attention of transformer can not handle modality alignment, and thus we add an alignment bias for audio-motion alignment. Third, we argue that modeling the correlation between speech and face motions needs to consider long-term audio context dependencies~\cite{karras2017audio}. Accordingly, we do not restrict the attention scope of the encoder self-attention, thus maintaining its ability to capture long-range audio context dependencies. Fourth, transformer with the sinusoidal position encoding has weak abilities to generalize to sequence lengths longer than the ones seen during training~\cite{press2021train,dehghani2018universal}. Inspired by Attention with Linear Biases (ALiBi)~\cite{press2021train}, we add a temporal bias to the query-key attention score and design a periodic positional encoding strategy to improve the model's generalization ability to longer audio sequences. 

The main contributions of our work are as follows:
\begin{itemize}
\item \textbf{An autoregressive transformer-based architecture for speech-driven 3D facial animation.} FaceFormer encodes the long-term audio context and the history of face motions to autoregressively predict a sequence of animated 3D face meshes. It achieves highly realistic and temporally stable animation of the whole face including both the upper face and the lower face. 

\item \textbf{The biased attention modules and a periodic position encoding strategy.} We carefully design the biased cross-modal MH attention to align the different modalities, and the biased causal MH self-attention with a periodic position encoding strategy to improve the generalization to longer audio sequences. 

\item \textbf{Effective utilization of the self-supervised pre-trained speech model.} Incorporating the self-supervised pre-trained speech model in our end-to-end architecture can not only handle the data limitation problem, but also notably improve the accuracy of mouth movements for the difficult cases, \eg, the lips are fully closed on /b/,/m/,/p/ phonemes.

\item \textbf{Extensive experiments and the user study to assess the quality of synthesized face motions.} The results demonstrate the superiority of FaceFormer over existing state-of-the-art methods in terms of realistic facial animation and lip sync on two 3D datasets~\cite{fanelli2010,cudeiro2019capture}.
\end{itemize}

\section{Related Work}

\subsection{Speech-Driven 3D Facial Animation}
Facial animation~\cite{weise2011realtime,li2013realtime,cao2016real,zollhofer2018state,kim2018deep,fried2019text,thies2020neural,lahiri2021lipsync3d} has attracted considerable attention over the years. While aware of extensive 2D-based approaches~\cite{fan2015photo,chung2016out,chen2018lip,prajwal2020lip,vougioukas2020realistic,das2020speech,yi2020audio,chen2020talking,ji2021audio,zhou2021pose}, we focus on animating a 3D model in this work. Typically, the procedural methods~\cite{massaro2012,taylor2012dynamic,xu2013practical,edwards2016jali} establish a set of explicit rules for animating the talking mouth. For example, the dominance functions~\cite{massaro2012} are used to characterize the speech control parameters. The dynamic viseme model proposed by Taylor \etal~\cite{taylor2012dynamic}  exploits the one-to-many mapping of phonemes to lip motions. Xu et al.~\cite{xu2013practical} construct a canonical set for modeling coarticulation effects. The state-of-the-art procedural approach JALI~\cite{edwards2016jali}
utilizes two anatomical actions to animate a 3D facial rig.

One appealing strength of the above procedural methods is the explicit control of the system to ensure the accuracy of the mouth movements. However, they require a lot of manual effort in parameter tuning. Alternatively, a wide variety of data-driven approaches~\cite{cao2005expressive,liu2015video,karras2017audio,taylor2017deep,pham2018end,cudeiro2019capture,hussen2020modality,richard2021meshtalk} has been proposed to produce 3D facial animation. Cao \etal~\cite{cao2005expressive} synthesize 3D facial animation based on the proposed Anime Graph structure and a search-based technique. The sliding window approach~\cite{taylor2017deep} requires the transcribed phoneme sequences as input and can re-target the output to other animation rigs. An end-to-end convolutional network elaborated by Karras \etal~\cite{karras2017audio} leverages the linear predictive coding method to encode audio and designs a latent code to disambiguate the variations in facial expression. Zhou \etal~\cite{zhou2018visemenet}
employ a three-stage network that combines phoneme groups, landmarks and audio features to predict viseme animation curves. VOCA~\cite{cudeiro2019capture} is a speaker-independent 3D facial animation method that captures a variety of speaking styles, yet the generated face motions are mostly present in the lower face. Recently, MeshTalk~\cite{richard2021meshtalk} learns a categorical latent space that successfully disentangles audio-correlated and audio-uncorrelated face motions. 

Most related to our work are methods~\cite{karras2017audio,cudeiro2019capture,richard2021meshtalk} whereby the high-resolution 3D data are used for training and the output is represented as the high-dimensional vector in 3D vertex space. The former two models~\cite{karras2017audio,cudeiro2019capture} are trained using short audio windows, thus ignoring the long-term audio context. Despite the highly realistic facial animation achieved the latter method~\cite{richard2021meshtalk}, it requires large amounts of high-fidelity 3D facial data to ensure the animation quality and the generalization to unseen identities.

\begin{figure*}
\centering
\includegraphics[width=0.7\textwidth]{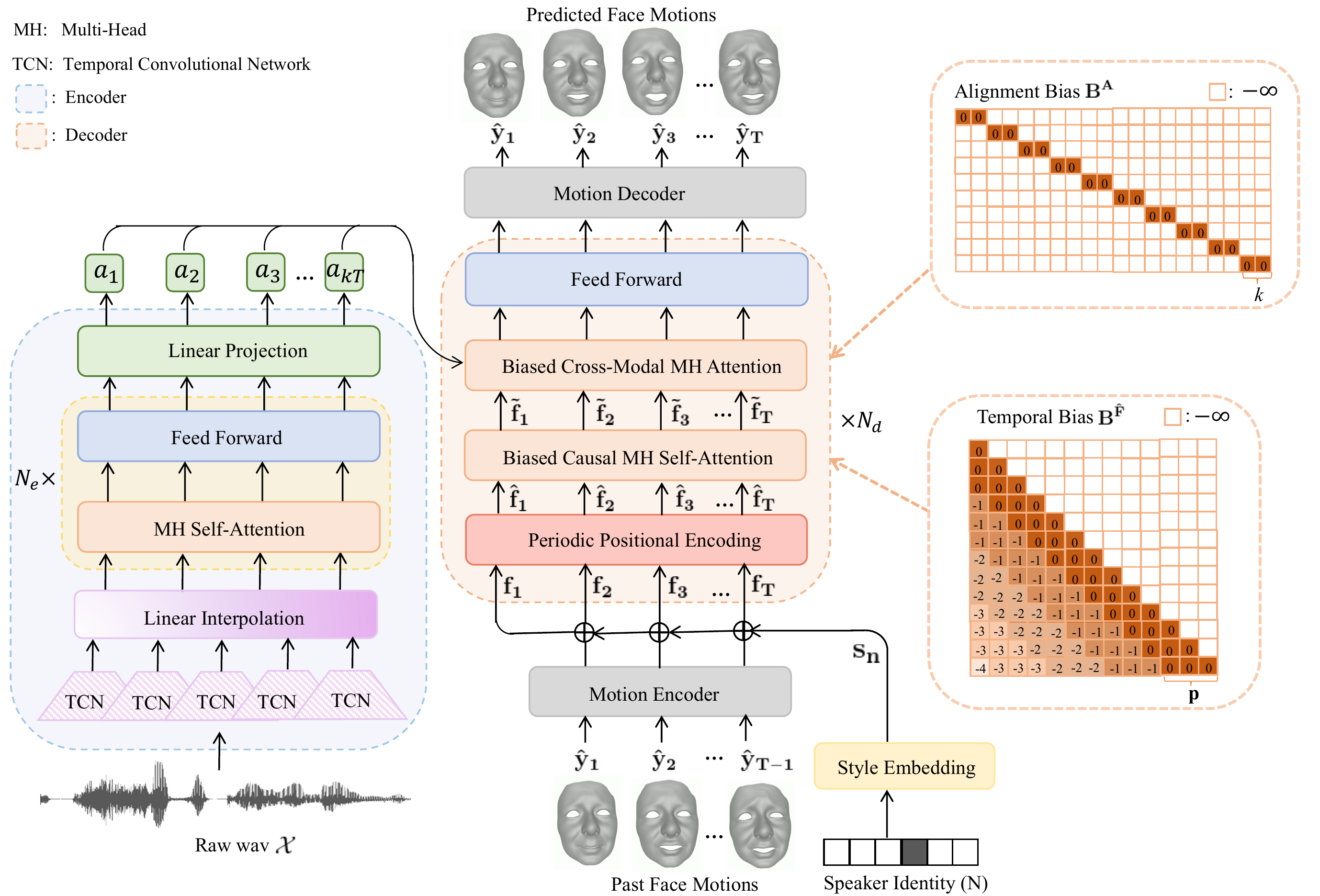}
\caption{\label{fig:overview} \textbf{Overview of FaceFormer.}
An encoder-decoder model with Transformer architecture takes raw audio as input and autoregressively generates a sequence of animated 3D face meshes. Layer normalizations and residual connections are omitted for simplicity. The overall design of the FaceFormer encoder follows wav2vec 2.0~\cite{baevski2020wav2vec}. In addition, a linear interpolation layer is added after TCN for resampling the audio features. We initialize the encoder with the corresponding pre-trained wav2vec 2.0 weights. The FaceFormer decoder consists of two main modules: a biased causal MH self-attention with a periodic positional encoding for generalizing to longer input sequences, and a biased cross-modal multi-head (MH) attention for aligning audio-motion modalities. During training, the parameters of TCN are fixed, whereas the other parts of the model are learnable.}
\end{figure*}

\subsection{Transformers in Vision and Graphics}

Transformer~\cite{vaswani2017attention} has emerged as a strong alternative to both RNN and CNN. In contrast to RNNs that process sequence tokens recursively, transformers can attend to all tokens in the input sequence parallelly, thereby modeling the long-range contextual information effectively. Vision Transformer (ViT)~\cite{dosovitskiy2020image} is the first work that explores the direct application of transformers to the task of image classification. Following ViT, some follow-up works~\cite{touvron2021training,chu2021conditional,chen2021crossvit} have been introduced to boost performance for image recognition problems. Besides, transformer-based models and the variants have also been proposed in object detection~\cite{carion2020end}, semantic segmentation~\cite{xie2021segformer}, image generation~\cite{jiang2021transgan}, \etc. In computer graphics, transformers have been exploited for 3D point cloud representations and 3D mesh, such as Point Transformer~\cite{zhao2020point}, Point Cloud Transformer~\cite{guo2021pct} and Mesh Transformer~\cite{lin2021end}. We refer readers to the comprehensive survey~\cite{khan2021transformers} for further information.

Some of the most recent works on 3D body motion synthesis~\cite{aksan2020spatio,petrovich2021action,bhattacharya2021text2gestures} and 3D dance generation~\cite{li2020learning,li2021AIST++,valle2021transflower} have explored the power of transformer in modeling sequential data and produced impressive results. Different from dance generation where the output motion is highly unconstrained, the task of speech-driven 3D facial animation inherently requires the alignment between audio and face motions to ensure the accuracy of lip motions. Meanwhile, the long-term audio context is expected to be considered, which is important for animating the whole face~\cite{karras2017audio}. Consequently, we present FaceFormer that incorporates the desirable properties for the speech-driven 3D facial animation problem.

\section{Our Approach: FaceFormer}

We formulate speech-driven 3D facial animation as
a sequence-to-sequence (seq2seq) learning problem and propose a novel seq2seq architecture (\cref{fig:overview}) to autoregressively predict facial movements conditioned on both audio context and past facial movement sequence. Suppose that there is a sequence of ground-truth 3D face movements $\mathbf{Y_{T}=(y_{1},...,y_{T}})$, where $\mathbf{T}$ is the number of visual frames, and the corresponding raw audio $\mathcal{X}$. The goal here is to produce a model that can synthesize facial movements $\mathbf{\hat{Y}_{T}}$ that is similar to $\mathbf{Y_{T}}$ given the raw audio $\mathcal{X}$. In the encoder-decoder framework (\cref{fig:overview}), the encoder first transforms $\mathcal{X}$ into speech representations $\mathbf{A_{T'}=(a_{1},...,a_{T'}})$, where $\mathbf{T'}$ is the frame length of speech representations. The style embedding layer contains a set of learnable embeddings that represents speaker identities $\mathbf{S=(s_{1},...,s_{N}})$. Then, the decoder autoregressively predicts facial movements $\mathbf{\hat{Y}_{T}=(\hat{y}_{1},...,\hat{y}_{T}})$ conditioned on $\mathbf{A_{T'}}$, the style embedding $\mathbf{s_n}$ of speaker $\mathbf{n}$, and the past facial movements. Formally,
\begin{equation} \label{eq-0}
 \mathbf{\hat{y}_{t}} =  \textbf{FaceFormer}_{\theta}(\mathbf{\hat{y}_{< t}, s_{n}}, \mathcal{X}),
\end{equation}
where $\theta$ denotes the model parameters, $\mathbf{t}$ is the current time-step in the sequence and $\mathbf{\hat{y}_{t}} \in \mathbf{\hat{Y}_{T}}$. For the remainder of this section, we describe each component of the FaceFormer architecture in detail.

\subsection{FaceFormer Encoder}

\subsubsection{\label{sec:wav2vec} Self-Supervised Pre-Trained Speech Model}
The design of our FaceFormer encoder follows the state-of-the-art self-supervised pre-trained speech model, wav2vec 2.0~\cite{baevski2020wav2vec}. Specifically, the encoder is composed of an audio feature extractor and a multi-layer transformer encoder~\cite{vaswani2017attention}. The audio feature extractor, which consists of several temporal convolutions layers (TCN), transforms the raw waveform input into feature vectors with frequency $f_a$. The transformer encoder is a stack of multi-head self-attention and feed-forward layers, converting the audio feature vectors into contextualized speech representations. The outputs of the temporal convolutions are discretized to a finite set of speech units via a quantization module. Similar to masked language modeling~\cite{devlin2018bert}, wav2vec 2.0 uses the context surrounding a masked time step to identify the true quantized speech unit by solving a contrastive task. 

We initialize our encoder (\cref{fig:overview}) with the pre-trained wav2vec 2.0 weights, and add a randomly initialized linear projection layer on the top. Since the facial motion data might be captured with a frequency $f_m$ that is different to $f_a$ (\eg, $f_a=49 Hz$ while for the BIWI datset~\cite{fanelli2010} $f_m=25 fps$), we add a linear interpolation layer after the temporal convolutions for resampling the audio features, which results in the output length $k\mathbf{T}$, where $k =\lceil \frac{f_a}{f_m} \rceil $. Therefore, the outputs of the linear projection layer can be represented as $\mathbf{A}_{k\mathbf{T}} = (\mathbf{a}_{1},...,\mathbf{a}_{k\mathbf{T}})$. In this way, the audio and motion modalities can be aligned by the biased cross-modal multi-head attention (\cref{sec:cross-modal attn}).

\subsection{FaceFormer Decoder}
\subsubsection{Periodic Positional Encoding}

 In practice, transformer has very limited generalization abilities for longer sequences due to the sinusoidal positional encoding method~\cite{dehghani2018universal,press2021train}. Attention with Linear Biases (ALiBi)~\cite{press2021train} method is proposed to improve generalization abilities by adding a constant bias to the query-key attention score. In our experiments, we notice that directly replacing the sinusoidal positional encoding with ALiBi would lead to a static facial expression during inference. This is because ALiBi does not add any position information to the input representation, which might influence the robustness of the temporal order information, especially for our case where the training sequences have subtle motion variations among adjacent frames. To alleviate this issue, we devise a periodic positional encoding (PPE) for injecting the temporal order information, while being compatible with ALiBi. Specifically, we modify the original sinusoidal positional encoding method~\cite{vaswani2017attention} to make it periodic with respect to a hyper-parameter $\mathbf{p}$ that indicates the period:
 \begin{equation}\label{eq-1}
\begin{aligned}
PPE_{(\mathbf{t}, 2\mathbf{i})} &=\sin \left( (\mathbf{t} \bmod \mathbf{p}) / 10000^{2\mathbf{i} / \mathbf{d}}\right) \\
PPE_{(\mathbf{t}, 2\mathbf{i}+1)} &=\cos \left( (\mathbf{t} \bmod \mathbf{p}) / 10000^{2\mathbf{i} / \mathbf{d}}\right)
\end{aligned}
\end{equation}
 where $\mathbf{t}$ denotes the token position or the current time-step in the sequence, $\mathbf{d}$ is the model dimension, and $\mathbf{i}$ is the dimension index. Rather than assigning a unique position identifier for each token~\cite{vaswani2017attention}, the proposed PPE strategy recurrently injects the position information within each period $\mathbf{p}$ (as shown in Section \ref{sec: Biased Causal atten}). Before PPE, we first project the face motion $\mathbf{\hat{y}_{t}}$ into a $\mathbf{d}$-dimensional space via a motion encoder. To model the speaking style, we embed the one-hot speaker identity to a $\mathbf{d}$-dimensional vector $\mathbf{s_{n}}$ via a style embedding layer and add it to the facial motion representation:
 \begin{equation}\label{eq-2}
\begin{aligned}
\mathbf{f_{t}}=\left\{\begin{array}{ll}
(\mathbf{W^{f}} \cdot \mathbf{\hat{y}_{t-1}}+\mathbf{b^{f}})+ \mathbf{s_{n}}, & 1 < \mathbf{t} \leq \mathbf{T}, \\
\mathbf{s_{n}}, &  \mathbf{t} = 1 ,
\end{array}\right.
\end{aligned}
\end{equation}
where $\mathbf{W^{f}}$ is the weight, $\mathbf{b^{f}}$ is the bias and $\mathbf{\hat{y}_{t-1}}$ is the prediction from the last time step. Then PPE is applied to $\mathbf{{f}_{t}}$ to provide the temporal order information periodically:
\begin{equation}\label{eq-3}
\mathbf{\hat{f}_{t}}= \mathbf{f_{t}} + PPE(\mathbf{t}).
\end{equation}
 

\subsubsection{Biased Causal Multi-Head Self-Attention}
\label{sec: Biased Causal atten}
 We design a biased causal multi-head (MH) self-attention mechanism based on ALiBi~\cite{press2021train}, which is reported to be beneficial for generalizing to longer sequences in language modeling. Given the temporally encoded facial motion representation sequence  $\mathbf{\hat{F}_{t}}=(\mathbf{\hat{f}_{1}},...,\mathbf{\hat{f}_{t})}$, biased causal MH self-attention first linearly projects $\mathbf{\hat{F}_{t}}$ into queries $\mathbf{Q^{\hat{F}}}$ and keys $\mathbf{K^{\hat{F}}}$ of dimension $\mathbf{d_k}$, and values $\mathbf{V^{\hat{F}}}$ of dimension $\mathbf{d_v}$. To learn the dependencies between each frame in the context of the past facial motion sequence, a weighted contextual representation is calculated by performing the scaled dot-product attention~\cite{vaswani2017attention}: \begin{equation}\label{eq-4}
\begin{aligned}
\operatorname{Att}(\mathbf{Q^{\hat{F}}},\!\mathbf{K^{\hat{F}}},\!\mathbf{V^{\hat{F}}},\!\mathbf{B^{\hat{F}}})\!=\!\operatorname{softmax}\!\left(\!\!\frac{\mathbf{Q^{\hat{F}}}\!\mathbf{(K^{\hat{F}})}^{T}}{\sqrt{\mathbf{d_k}}}\!\!+\!\!\mathbf{B^{\hat{F}}}\!\!\right)\!\mathbf{V^{\hat{F}}},
\end{aligned}
\end{equation}
where $\mathbf{B^{\hat{F}}}$ is the temporal bias we add to ensure causality and to improve the ability to generalize to longer sequences. 

More specifically, $\mathbf{B^{\hat{F}}}$ is a matrix that has negative infinity in the upper triangle to avoid looking at future frames to make current predictions. For the generalization ability, we add static and non-learned biases to the lower triangle of $\mathbf{B^{\hat{F}}}$. Different from ALiBi~\cite{press2021train}, we introduce the period $\mathbf{p}$ and inject the temporal bias to each period $([1:\mathbf{p}], [\mathbf{p}+1:\mathbf{2p}], \dots)$. Let us define $i$ and $j$ as the indices of $\mathbf{B^{\hat{F}}}$ ($1 \leq i \leq \mathbf{t}$, $1 \leq j \leq \mathbf{t}$). Then the temporal bias $\mathbf{B^{\hat{F}}}$ is formulated as:
\begin{equation}\label{eq-5}
\begin{aligned}
\mathbf{B^{\hat{F}}}\left(i,j\right)=\left\{\begin{array}{ll}
\lfloor (i - j)/\mathbf{p} \rfloor, &  j \leq i ,\\
-\infty, & \text{otherwise}.
\end{array}\right.
\end{aligned}
\end{equation}
In this way, we bias the casual attention by assigning higher attention weights to the closer period. Intuitively, the closest facial frames period $\mathbf{(\hat{y}_{t-\mathbf{p}},...,\hat{y}_{t-1}})$ are most likely to affect the current prediction of $\mathbf{\hat{y}_{t}}$. Thus, our proposed temporal bias can be considered as a generalized form of ALiBi and ALiBi becomes a special case when $\mathbf{p}=1$.

The MH attention mechanism, which consists of $\mathbf{H}$ parallel scaled dot-product attentions, is applied to jointly extract the complementary information from multiple representation subspaces. The outputs of $\mathbf{H}$ heads are concatenated together and projected forward by a parameter matrix $\mathbf{W^{\hat{F}}}$:
\begin{equation}\label{eq-6}
\begin{aligned}
\operatorname{MH}(\mathbf{Q^{\hat{F}}}\!,\!\mathbf{K^{\hat{F}}}\!,\!\mathbf{V^{\hat{F}}}\!,\!\mathbf{B^{\hat{F}}}) &=\operatorname{Concat}\left(\operatorname{head}_{1},\ldots,\!\text{head}_{\mathbf{H}}\right)\!\mathbf{W^{\hat{F}}},\\
\text { where head }_{\mathbf{h}} &=\operatorname{Att}\left(\mathbf{Q^{\hat{F}}_{h}}, \mathbf{K^{\hat{F}}_{h}}, \mathbf{V^{\hat{F}}_{h}}, \mathbf{B^{\hat{F}}_{h}}\right).
\end{aligned}
\end{equation}

Similar to ALiBi~\cite{press2021train}, we add a head-specific scalar $\mathbf{m}$ for the MH setting. For each $\text{head}_{\mathbf{h}}$, the temporal bias is defined as $\mathbf{B^{\hat{F}}_{h}}$ = $\mathbf{B^{\hat{F}}}\cdot \mathbf{m}$. The scalar $\mathbf{m}$ is a head-specific slope and is not learned during training. For $\mathbf{H}$ heads, $\mathbf{m}$ will start at $2^{-2^{\left(-\log _{2} \mathbf{H}+3\right)}}$ and multiply each element by the same value to compute the next element. Concretely, if the model has 4 heads, the corresponding slopes will be $2^{-2}$, $2^{-4}$, $2^{-6}$ and $2^{-8}$.

\subsubsection{Biased Cross-Modal Multi-Head Attention}
\label{sec:cross-modal attn}

The biased cross-modal multi-head attention aims to combine the outputs of Faceformer encoder (speech features) and biased causal MH self-attention (motion features) to align the audio and motion modalities (see \cref{fig:overview}). For this purpose, we add an alignment bias to the query-key attention score, which is simple and effective. The alignment bias $\mathbf{B^{A}}$ ($1 \leq i \leq  \mathbf{t} , 1 \leq j \leq k\mathbf{T}$) is represented as:
 \begin{equation}\label{eq-7}
\mathbf{B^{A}}(i, j)=\left\{\begin{array}{ll}
0, & ki \leq j < k(i+1) \\
-\infty, & \text { otherwise }
\end{array},\right.
 \end{equation}

 Each token in $\mathbf{{A}}_{k\mathbf{T}}$ has captured the long-term audio context due to the self-attention mechanism. On the other hand, assuming the outputs of biased causal MH self-attention is $\mathbf{\tilde{F}_{t}=(\tilde{f}_{1},...,\tilde{f}_{t})}$, each token in $\mathbf{\tilde{F}_{t}}$ has encoded the history context of face motions. Both $\mathbf{{A}}_{k\mathbf{T}}$ and $\mathbf{\tilde{F}_{t}}$ are fed into biased cross-modal MH attention. Likewise, $\mathbf{{A}}_{k\mathbf{T}}$ is transformed into two separate matrices: keys $\mathbf{{K}^{A}}$ and values $\mathbf{{V}^{A}}$, whereas $\mathbf{\tilde{F}_{t}}$ is transformed into queries $\mathbf{{Q}^{\tilde{F}}}$. The output is calculated as a weighted sum of $\mathbf{{V}^{A}}$, 
\begin{equation}\label{eq-8}
\begin{aligned}
\operatorname{Att}(\mathbf{Q^{\tilde{F}}}\!,\!\mathbf{K^{A}}\!,\!\mathbf{V^{A}}\!,\!\mathbf{B^{A}})=\operatorname{softmax}\!\left(\!\!\frac{\mathbf{Q^{\tilde{F}}}\!\mathbf{(K^{A})}^{T}}{\sqrt{\mathbf{d_k}}}\!\!+\!\!\mathbf{B^{A}}\!\!\right)\!\mathbf{V^{A}}.
\end{aligned}
\end{equation}

To explore different subspaces, we also extend \cref{eq-8} to $\mathbf{H}$ heads as in \cref{eq-6}. Finally, the predicted face motion $\mathbf{\hat{y}_{t}}$ is obtained by projecting the $\mathbf{d}$-dimensional hidden state back to the $\mathbf{V}$-dimensional 3D vertex space via a motion decoder.

\subsection{Training and Testing}
During the training phase, we adopt an autoregressive scheme instead of a teacher-forcing scheme. In our experiments, we observe that training FaceFormer with a less guided scheme works better than a fully guided one. Once the complete 3D facial motion sequence is produced, the model is trained by minimizing the Mean Squared Error (MSE) between the decoder outputs $\mathbf{\hat{Y}_{t}}=(\mathbf{\hat{y}_{1}},...,\mathbf{\hat{y}_{T})}$ and the ground truth $\mathbf{{Y}_{t}}=(\mathbf{{y}_{1}},...,\mathbf{{y}_{T}})$:

\begin{equation}\label{eq-9}
\mathcal{L}_{\text{MSE}}=\sum_{\mathbf{t}=1}^{\mathbf{T}} \sum_{\mathbf{v}=1}^{\mathbf{V}}\left\|\hat{\mathbf{y}}_{\mathbf{t},\mathbf{ v}}-\mathbf{y}_{\mathbf{t}, \mathbf{v}}\right\|^{2},
\end{equation}
where $\mathbf{V}$ represents the number of vertices of the 3D face mesh. 

At inference time, FaceFormer autoregressively predicts a sequence of animated 3D face meshes. More specifically, at each time-step,
it predicts the face motion $\mathbf{\hat{y}_{t}}$ conditioned on the raw audio $\mathcal{X}$, the history of face motions $\mathbf{\hat{y}_{<t}}$ and the style representations $\mathbf{s_n}$ as in \cref{eq-0}. $\mathbf{s_n}$ is determined by the speaker identity, and thus altering the one-hot identity vector can manipulate the output in different styles.

\section{Experiments and Results}

\subsection{Experimental Settings}

We use two publicly available 3D datasets, BIWI~\cite{fanelli2010} and VOCASET~\cite{cudeiro2019capture} for training and testing. Both datasets provide the audio-3D scan pairs of English spoken utterances. BIWI contains 40 unique sentences shared across all speakers. VOCASET contains 255 unique sentences, some of which are shared across speakers. Comparatively, BIWI represents a more challenging dataset for lip sync as it covers fewer phonemes.

\textbf{BIWI Dataset.} BIWI is a corpus of affective speech and corresponding dense dynamic 3D face geometries. 14 human subjects are asked to read 40 English sentences, each of which is recorded twice: in a neutral or emotional context. The 3D face geometries are captured at 25fps, each with 23370 vertices. Each sequence is 4.67 seconds long on average. For our experiments, we use the subset where the sentences are recorded in the emotional context. Specifically, we split the data into a training set (BIWI-Train) of 192 sentences spoken by six subjects (each subject speaks 32 sentences), a validation set (BIWI-Val) of 24 sentences spoken by six subjects (each subject speaks 4 sentences), and two testing sets (BIWI-Test-A and BIWI-Test-B). BIWI-Test-A includes 24 sentences spoken by six seen subjects (each speaks 4 sentences), and BIWI-Test-B includes 32 sentences spoken by eight unseen subjects (each speaks 4 sentences). 

\textbf{VOCASET Dataset.} VOCASET is composed of 480 facial motion sequences from 12 subjects. Each sequence is captured at 60fps and is between 3 and 4 seconds long. Each 3D face mesh has 5023 vertices. For a fair comparison, we use the same training, validation and testing splits as VOCA~\cite{cudeiro2019capture}, which we refer to VOCA-Train, VOCA-Val and VOCA-Test, respectively.

\textbf{Baseline Methods.} We compare FaceFormer with two state-of-the-art methods, VOCA~\cite{cudeiro2019capture} and MeshTalk~\cite{richard2021meshtalk}, on both BIWI and VOCASET. Among the three methods, FaceFormer and VOCA require conditioning on a training speaker identity during inference. For unseen subjects, we obtain the predictions of FaceFormer and VOCA by conditioning on all training identities. The implementation details of FaceFormer and the baseline methods are provided in the supplementary material (Sec. 1 and Sec. 2).

\subsection{Evaluation Results}

\begin{table}
\centering
\vspace{-0.1in}
 \caption{\label{tb:compare} 
Comparison of lip-sync errors. We compare FaceFormer with two state-of-the-art methods~\cite{cudeiro2019capture,richard2021meshtalk} on BIWI-Test-A. The average lip error~\cite{richard2021meshtalk} is used for lip synchronization evaluation.} 
\resizebox{0.33\textwidth}{!}{
 \begin{tabular}{lcc}
 \toprule
\multicolumn{2}{l|}{\textbf{Methods}}& \textbf{Lip Vertex Error ($\times10^{-4}$mm)}  \\
\midrule 
\multicolumn{2}{l|}{VOCA}& 7.6427	\\
\multicolumn{2}{l|}{MeshTalk}& 6.7436\\
\multicolumn{2}{l|}{FaceFormer (Ours)}& \textbf{5.3742} \\
\bottomrule
 \end{tabular} 
 }
\end{table}

\begin{table}
\centering
\vspace{-0.1in}
 \caption{\label{tb:AMT-BIWI} User study results on BIWI-Test-B. We use A/B testing and report the percentage of answers where A is preferred over B.}
 \resizebox{0.4\textwidth}{!}{
 \begin{tabular}{lccc}
 \toprule
\multicolumn{2}{l|}{\textbf{Ours \vs Competitor}}&\textbf{Realism} &\textbf{Lip Sync}   \\
\midrule 
\multicolumn{2}{l|}{Ours \vs VOCA}& $83.85\pm3.76$  & $82.64\pm3.77$   \\
\multicolumn{2}{l|}{Ours \vs MeshTalk}& $83.33\pm4.07$  & $80.56\pm5.22$ \\
\multicolumn{2}{l|}{Ours \vs GT}&  $35.24\pm2.87$  & $36.98\pm1.38$   \\
\bottomrule
 \end{tabular} 
 }
\end{table}

\begin{table}
\centering
\vspace{-0.1in}
 \caption{\label{tb:AMT-VOCA} User study results on VOCA-Test.} 
 \resizebox{0.4\textwidth}{!}{
 \begin{tabular}{lccc}
 \toprule
\multicolumn{2}{l|}{\textbf{Ours \vs Competitor}}&\textbf{Realism} &\textbf{Lip Sync}   \\
\midrule 
\multicolumn{2}{l|}{Ours \vs VOCA}& $77.92\pm7.94$ & $77.08\pm7.32$\\
\multicolumn{2}{l|}{Ours \vs MeshTalk}& $82.92\pm2.60$ &  $82.08\pm3.15$\\
\multicolumn{2}{l|}{Ours \vs GT}& $29.17\pm10.41$  &$30.42\pm8.04$  \\
\bottomrule
 \end{tabular}
 }
\end{table}

\textbf{Lip-sync Evaluation.} We follow the lip-sync  metric employed in MeshTalk~\cite{richard2021meshtalk} for evaluating the quality of lip movements. The maximal L2 error of all lip vertices is defined as the lip error for each frame. The error is calculated by comparing the predictions and the captured 3D face geometry data. We report the computed average over all testing sequences of BIWI-test-A
for VOCA, MeshTalk and FaceFormer in \cref{tb:compare}. The lower average lip error achieved by FaceFormer suggests it can produce more accurate lip movements compared to the other two methods.

\textbf{Qualitative Evaluation.} Given the many-to-many mappings between upper face motions and the speech utterance, it is suggested that qualitative evaluations and user studies are more proper for evaluating the quality of speech-driven facial animation than using quantitative metrics~\cite{cudeiro2019capture,karras2017audio}. We refer the readers to our supplementary video for the assessment of the motion quality. The video compares the results of our approach, those by the previous methods~\cite{taylor2017deep,karras2017audio,cudeiro2019capture,richard2021meshtalk} and the ground truth. Specifically, we test our model using (1) audio sequences from BIWI and VOCASET test sets, (2) audio clips extracted from supplementary videos of previous methods and (3) audio clips extracted from TED videos on YouTube. For the last two cases, the results are predicted from the model trained on BIWI. The video shows that FaceFormer produces realistic and natural-looking facial animation with accurate lip synchronization. Compared to VOCA and MeshTalk, it is notable that, FaceFormer produces more realistic facial motions and better lip sync with proper mouth closures in many situations, \eg, the lips are fully closed when pronouncing /b/,/m/,/p/. 
We also show that our system can produce animation of talking in different styles and different languages. 

\subsection{Perceptual Evaluation}
\label{sec:perceptual}

\textbf{User Study on BIWI.} We conduct user studies on Amazon Mechanical Turk (AMT) to evaluate the animation quality of FaceFormer, compared with the ground truth, VOCA and MeshTalk. For BIWI, we obtain the results of three methods using all test audio sequences of BIWI-Test-B (32 sentences). The results of FaceFormer and VOCA are produced by conditioning on all training speaker identities, which results in 192 videos (32 sentences $\times$ 6 identities) for each method. Therefore, 576 A \vs B pairs (192 videos $\times$ 3 comparisons) are created for BIWI-Test-B. For each HIT (human intelligence task), the AMT interface shows four video pairs including the qualification test in randomized order, and the Turker is instructed to judge the videos in terms of realistic facial animation and lip sync. Each video pair is evaluated by three Turkers. In particular, Turkers must pass the qualification test otherwise they are not allowed to submit HITs. Finally, we collect 576 HITs for the user study on BIWI. More details about the user study are described in our supplementary material (Sec. 3).

\cref{tb:AMT-BIWI} shows the percentage of A/B testing in terms of realism and lip sync. Turkers favor FaceFormer over VOCA in terms of realistic facial animation and lip sync. We believe this is mainly due to two reasons: (1) face motions synthesized by VOCA are mostly present in the lower face; (2) VOCA sometimes fails to fully close the mouth at the phonemes /b/,/m/,/p/. FaceFormer also outperforms MeshTalk and we attribute this to the results produced by FaceFormer having more expressive facial motions and more accurate mouth movements. Not surprisingly, Turkers perceive the ground truth more realistic than FaceFormer. 

\textbf{User Study on VOCASET.} In the second user study, we compare the results of three methods on VOCA-Test. We randomly select 10 sentences from VOCA-Test and obtain the results of FaceFormer and VOCA conditioned on all training speaker identities, which results in 80 videos (10 sentences $\times$ 8 identities) for each method. In total, 240 A \vs B pairs (80 videos $\times$ 3 comparisons) are created for VOCA-Test. Similarly, for each pair, Turkers make the choice between two videos in terms of realism and lip sync. Since VOCASET has very few upper face motions, movements are present mostly in the lower face for all three methods. In this case, well-synchronized lip motions are important for generating perceptually realistic results. \cref{tb:AMT-VOCA} shows that FaceFormer achieves higher percentages over VOCA and MeshTalk. We believe this is because our results have better synchronized mouth shapes and closures. Similarly, there is still a certain gap between our results and the ground truth.

\subsection{Visualization Analysis}
\label{sec:visual}

\begin{figure}
\centering
\includegraphics[width=0.46\textwidth]{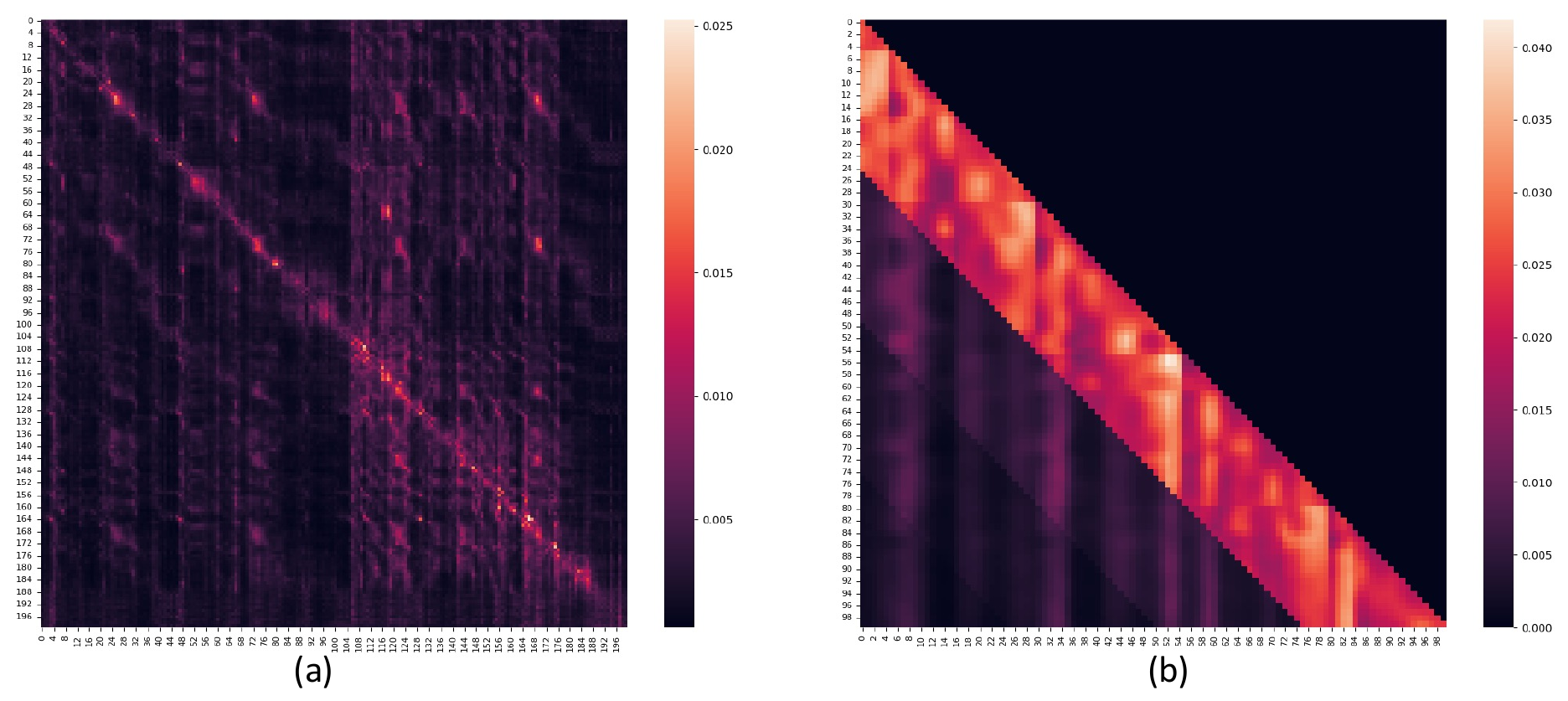}
\vspace{-0.1in}
\caption{\label{fig:attention} \textbf{Attention Weights Visualization.} Attention weights of the (a) MH self-attention of the encoder and (b) biased causal MH self-attention of the decoder.}
\end{figure}

To provide insights into the underlying attention mechanism, we visualize the attention weights for the MH self-attention of the encoder, as well as the biased causal MH self-attention of the decoder. We consider 100 frames of a test sequence from BIWI and examine the attention weights that are used to predict the last frame. \cref{fig:attention} visualizes the average attention weights across all heads. We observe that the encoder self-attention (\cref{fig:attention} (a)) not only focuses on the nearby audio frames (as reflected by the diagonal line) but also attends to some farther future and past frames. This indicates that the self-attention mechanism of transformer is able to capture both the short- and long-range audio context dependencies.
The attended audio frames may contain more informative context features that influence the current face motion. For the decoder self-attention (\cref{fig:attention} (b)), the visualization corresponds to the casual attention incorporated with the temporal bias (\cref{eq-5}). There is a clear pattern that the face motion frames in a closer period are assigned with higher weights, as those frames are more likely to influence the current face motion. For example, there is a high probability that people will keep smiling if they have been smiling over the past frames.

\subsection{Ablation Study}  
\label{sec:ablation}

The visual results of ablation study are included in the supplementary video. Please watch the supplementary video for the dynamic comparison. 

\subsubsection{Ablation on FaceFormer Encoder} 

\hspace{0.1in} \textbf{Effect of the encoder self-attention module.} To investigate the effect of the MH self-attention module in FaceFormer encoder, we directly remove it from the whole architecture, with the pre-trained TCN retained to extract the speech representations. We refer to this variant as ``TCN+FaceFormer Decoder'' and conduct the comparison experiments on BIWI. The results show that ``TCN+FaceFormer Decoder'' often fails to close the mouth, resulting in out-of-sync lip motions. Besides, the produced results have a temporal jitter effect around the mouth region, as shown in the supplementary video.
 
\textbf{Effect of the wav2vec weights initialization.} We also perform an ablation study of the wav2vec weights initialization by comparing FaceFormer trained with and without wav2vec weights initialization (denoted as ``FaceFormer w/o wav2vec''). Without wav2vec weights initialization, we observe a downgrade of the quality of face movements. ``FaceFormer w/o wav2vec'' can not produce synchronized mouth motions and a temporal jitter effect can be observed. This suggests that simply training FaceFormer with randomly initialized weights might converge to a poor solution. Hence, the wav2vec weights initialization is necessary for the FaceFormer encoder.

\subsubsection{Ablation on FaceFormer Decoder}  

\hspace{0.1in} \textbf{Choices of the decoder architecture.} We explore whether the transformer-based architecture has advantages over a fully-connected layer or LSTM by training and testing two alternative variants: ``FaceFormer Encoder+FC'' and ``FaceFormer Encoder+LSTM''. As shown in the supplemental video, FaceFormer yields more stable mouth motions and more accurate lip sync compared to the two variants. Compared to the FC decoder, the autoregressive machanism of FaceFormer decoder can stabilize the predicted lip motions by modeling the history motions. On the other hand, the self-attention machanism of FaceFormer decoder might model the context cues in history motions better than LSTM, thus having more temporally coherent lip motions.

\textbf{Effect of the alignment bias.} We examine the effect of the alignment bias (\cref{eq-7}) by removing it from the biased cross-modal MH attention module. The model without the alignment bias (denoted as ``FaceFormer w/o AB'') tends to generate muted facial expressions across all frames. Hence, the alignment bias is indispensable for the cross-modal attention in aligning the audio-motion modalities correctly.

\begin{figure}
\centering
\includegraphics[width=0.48\textwidth]{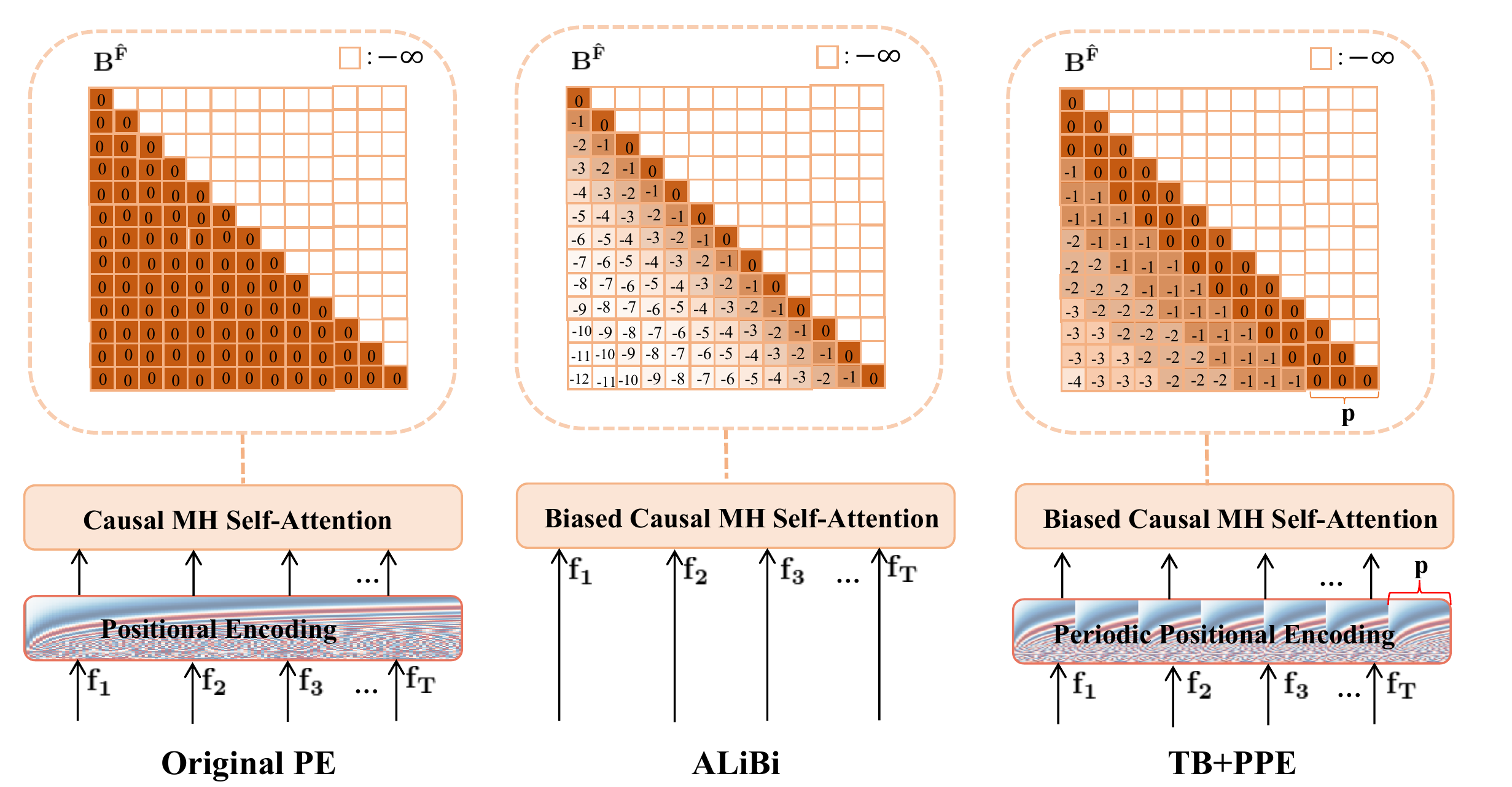}
\vspace{-0.2in}
\caption{\label{fig:difference} Illustration of different positional encoding strategies. }
\end{figure}

\textbf{Effect of the proposed positional encoding strategy.} 
In the FaceFormer decoder, the proposed positional encoding strategy is adding a temporal bias to the attention score and making the original sinusoidal position embedding~\cite{vaswani2017attention} periodic. We refer to this strategy as ``TB+PPE''. We compare ``TB+PPE'' with the original sinusoidal position encoding~\cite{vaswani2017attention} (``Original PE'') and ``ALiBi''~\cite{press2021train}. The differences of three different positional encoding strategies are visualized in \cref{fig:difference}. The results show ``Original PE'' can still produce well-synchronized mouth motions with proper lip closures, yet has a temporal jitter effect around the lips during silent frames, especially as the test audio sequence exceeds the average length of training audio sequences. While ``ALiBi'' does not influence the results on BIWI, it quickly freezes to a static facial expression when training and testing on VOCASET. This happens because the original ALiBi does not add any position information to the input representation, influencing the robustness of the temporal order information. This influence is more obvious when motion data have subtle variations among adjacent frames.


Since ``TB+PPE'' is a key component for improving the ability to generalize to longer audio sequences, we additionally study its influence by conducting the perceptual evaluation on AMT. Specifically, we download the TED videos shared under the ``CC BY-NC-ND 4.0 International License'', and extract 15 representative audio clips for the user study. The audio sequences are around 20 seconds long, more than four times the average length of training audio sequences. For the comparison to ``Original PE'', we randomly sample a training identity and use it as the condition for both methods. Similar to the user study in \cref{sec:perceptual}, each video pair is evaluated by three judges. Overall, Turkers perceive the facial animation results of FaceFormer more realistic ($57.78\% \pm 16.78\%$) and the generated lip motions of FaceFormer more in sync with audio ($62.22\% \pm 10.18\%$) than ``Original PE''. That indicates that FaceFormer generalizes better to longer audio clips than ``Original PE''. The likely explanation is that ``Original PE'' tends to generate unstable lip motions during silent frames when testing on longer audio sequences.

\section{Discussion and Conclusion}

In this work, we propose an autoregressive transformer-based architecture for speech-driven 3D facial animation. The encoder effectively leverages the self-supervised pre-trained speech representations, and the inside self-attention can capture long-range audio context dependencies. The decoder attention modules with a periodic position encoding strategy are tailored for cross-modal alignment and generalization to longer sequences. Overall, FaceFormer demonstrates higher quality for lip synchronization and realistic facial animation compared to the state-of-the-arts. However, the main bottleneck in our model is the quadratic memory and time complexity of the self-attention mechanism, making it not suitable for real-time applications. One future work is to address this problem using advanced techniques~\cite{wang2020linformer,beltagy2020longformer} that improve the efficiency of self-attention. 

\noindent
{\bf Ethics Considerations:}
We should use technology responsibly and be careful about the synthesized content. Since our technique requires 3D scan data collected from actors, it is important to obtain consent from the actors during data acquisition. Our method can animate a realistic 3D talking face from an arbitrary audio signal. However, there is a risk that such techniques could potentially be misused to cause embarrassment. Thus, we hope to raise the public’s awareness about the risks of the potential misuse and encourage research efforts on the responsible use of technology.

\noindent
{\bf Acknowledgement.}
This research is partly supported by New Energy and Industrial Technology Development Organization (NEDO) (ref:JPNP21004).

{\small
\bibliographystyle{ieee_fullname}
\bibliography{faceformer}
}

\clearpage

\section*{\Large Supplementary Material}

In this supplementary material, we provide further information about FaceFormer, including detailed explanation of the FaceFormer architecture and the training details (\cref{sec:implementation}), implementations of baseline methods (\cref{sec:comparing}), and the additional information about user study (\cref{sec:user}).

\section{Implementation Details}
\label{sec:implementation}

\textbf{Network architecture.} The overall architecture of FaceFormer is illustrated in Fig. 2 of the main paper. In the encoder, the TCN is followed by a linear interpolation layer, which down/up samples the input to a given size determined by the frequency of the captured facial motion data. The interpolated outputs are then fed into 12 identical transformer encoder layers. For each transformer encoder layer, the model dimensionality is 768 and the number of attention heads is 12. Next, a linear projection layer is added on top of the transformer encoder layers, converting the 768-dimensional features to $\mathbf{d}$-dimensional speech representations ($\mathbf{d}$ = 128 for BIWI and $\mathbf{d}$ = 64 for VOCASET).

The motion encoder is a fully-connected layer with $\mathbf{d}$ outputs and the style embedding layer is an embedding layer with $\mathbf{d}$ outputs. The FaceFormer decoder has one decoder layer. The periodic positional encodings (PPE) have the same dimension as the motion encoder so that the two can be summed. For both the biased causal MH self-attention and the biased cross-modal MH attention, we employ 4 heads and set the model dimensionality to $\mathbf{d}$. The dimension of the FF layer is 2048. Similar to the encoder, the residual connections and layer normalizations are applied to the two biased attention layers and the FF layer. Finally, a fully-connected layer with $\mathbf{v}$ outputs is applied as the motion decoder ($\mathbf{v}$ = 70110 for BIWI and $\mathbf{v}$ = 15069 for VOCASET).

\textbf{Training.} We use the Adam optimizer with a learning rate of 1e-4. The parameters of the encoder are initialized with the pre-trained wav2vec 2.0 weights. During training, only the parameters of TCN are fixed. The models are trained for 100 epochs. The period $\mathbf{p}$ is set to 25 for BIWI and 30 for VOCASET.

\section{Baseline Methods}
\label{sec:comparing}
As mentioned in the main paper, we compare FaceFormer with two state-of-the-art methods, VOCA~\cite{cudeiro2019capture} and MeshTalk~\cite{richard2021meshtalk}, on both the BIWI and VOCASET datasets. For comparisons on BIWI, we use the original implementation of VOCA from their codebase\footnote{https://github.com/TimoBolkart/voca}. Specifically, we train and test the VOCA model on BIWI. For comparisons on VOCASET, we directly use the provided trained VOCA model from their codebase and test it on VOCA-Test. Besides, we implement MeshTalk to the best of our understanding. We train and test MeshTalk on BIWI and VOCASET, respectively. One difference is the design of the UNet-style decoder: We modified the number of fully-connected layers from 7 to 3, as we found the original decoder would lead to overfitting due to the limited number of identities in BIWI and VOCASET. 

\begin{figure*}[!htbp]
\centering        
    \includegraphics[width=1.9\columnwidth]{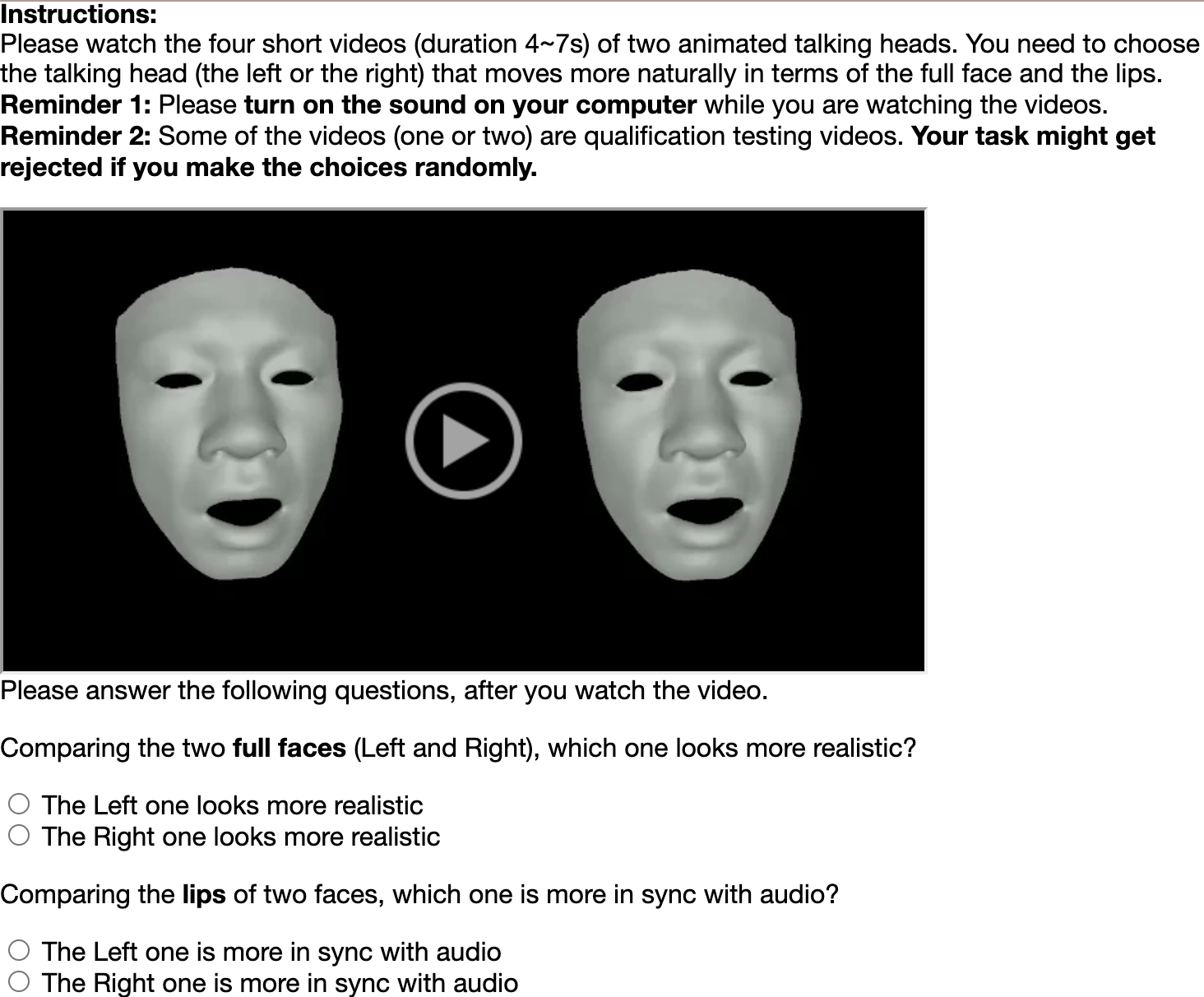}  
    \caption{\label{fig:AMT1}Designed user interface on AMT. Each HIT contains four video pairs and here only one video pair is shown due to the page limit.}
\end{figure*}

\begin{figure*}[!htbp]
\centering        
    \includegraphics[width=1.9\columnwidth]{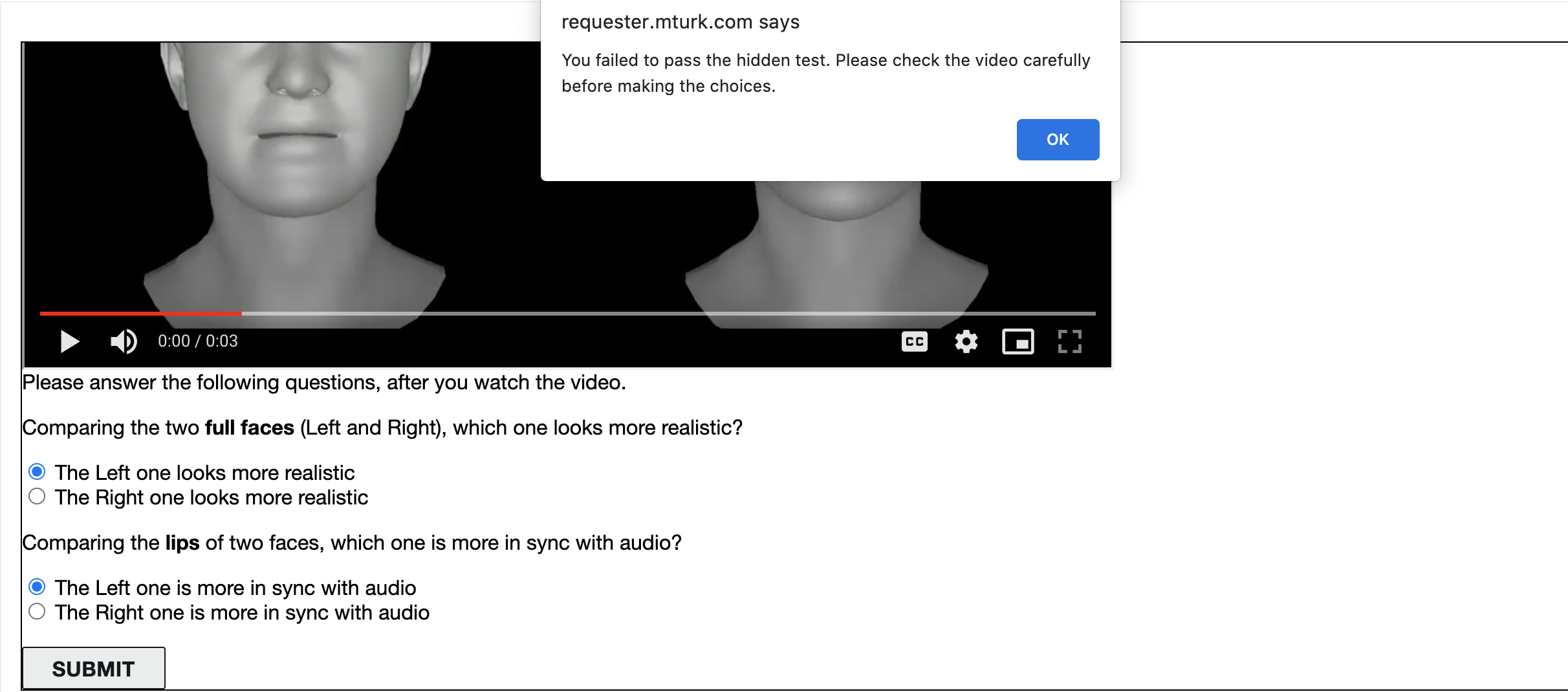}  
    \caption{\label{fig:AMT2}Screen shot of the warning message.}
\end{figure*}

\section{User Study}
\label{sec:user}
The designed user interface on Amazon Mechanical Turk (AMT) is shown in \cref{fig:AMT1}. To avoid Turkers' selecting an option randomly, we add one or two qualification testing videos for each HIT (human intelligence task). As shown in \cref{fig:AMT2}, Turkers could not submit their answers successfully if they failed to pass the hidden test. A warning message would pop up asking for checking the videos carefully before making the choices. Our recruitment requirement is that the Turkers have finished over 5000 HITs before and have an approval rate of at least 98$\%$. In total, 576 A \vs B pairs (192 videos $\times$ 3 comparisons) are created for BIWI-Test-B, and 240 A \vs B pairs (80 videos $\times$ 3 comparisons) are created for VOCA-Test. For each HIT (human intelligence task), the AMT interface shows four video pairs including the qualification test in randomized order, and the Turker is instructed to judge the videos w.r.t two questions: ``Comparing the two full faces, which one looks more realistic?'' and ``Comparing the lips of two faces, which one is more in sync with audio?''.

\end{document}